\documentclass[a4paper, 10pt, conference]{ieeeconf}      
\usepackage{FG2020}
\usepackage[ruled]{algorithm2e}
\usepackage{balance}
\usepackage{float}
\usepackage{amsmath,epsfig}
\usepackage{tabularx,mathrsfs,graphicx,epstopdf,multirow,epsfig,cite,float,caption,booktabs,makecell,multirow,array,amsfonts,setspace,pifont,amssymb,setspace,algorithmicx,algpseudocode,bm}
\usepackage[colorlinks,linkcolor=red,anchorcolor=blue,citecolor=green]{hyperref}

\FGfinalcopy 

\IEEEoverridecommandlockouts                              
\def\FGPaperID{***} 

\title{\LARGE \bf
Deep Fusion Siamese Network for Automatic Kinship Verification
}
\author{\parbox{16cm}{\centering
    {\large Jun Yu, Mengyan Li{$^\dag$}, Xinlong Hao and Guochen Xie}\\
	 {harryjun@ustc.edu.cn, \{limmy, haoxl, xiegc\}@mail.ustc.edu.cn}\\
    {\normalsize Department of Automation, University of Science and Technology of China, Hefei, CN }
}
    \thanks{{$^\dag$}Corresponding author of this paper. }
}

\begin{document}

\ifFGfinal
\thispagestyle{empty}
\pagestyle{empty}
\else
\author{Anonymous FG2020 submission\\ Paper ID \FGPaperID \\}
\pagestyle{plain}
\fi
\maketitle

\begin{abstract}
Automatic kinship verification aims to determine whether some individuals belong to the same family. It is of great research significance to help missing persons reunite with their families. In this work, the challenging problem is progressively addressed in two respects. First, we propose a deep siamese network to quantify the relative similarity between two individuals. When given two input face images, the deep siamese network extracts the features from them and fuses these features by combining and concatenating. Then, the fused features are fed into a fully-connected network to obtain the similarity score between two faces, which is used to verify the kinship. To improve the performance, a jury system is also employed for multi-model fusion. Second, two deep siamese networks are integrated into a deep triplet network for tri-subject (i.e., father, mother and child) kinship verification, which is intended to decide whether a child is related to a pair of parents or not. Specifically, the obtained similarity scores of father-child and mother-child are weighted to generate the parent-child similarity score for kinship verification. Recognizing Families In the Wild (RFIW) is a challenging kinship recognition task with multiple tracks, which is based on Families in the Wild (FIW), a large-scale and comprehensive image database for automatic kinship recognition. The Kinship Verification (track I) and Tri-Subject Verification (track II) are supported during the ongoing RFIW2020 Challenge. Our team (ustc-nelslip) ranked 1st in track II, and 3rd in track I. 
The code is available at {\color{blue}https://github.com/gniknoil/FG2020-kinship}.

\end{abstract}

\section{INTRODUCTION}

Automatic kinship verification, as a classical Boolean problem, is used to predict whether given face images have kin relations. It is essential in various real-world applications such as genealogical studies \cite{c1}, social-media analysis \cite{c2}, and tracking missing persons \cite{c3}.

\begin{figure}[t]
\setlength{\abovecaptionskip}{0.0cm}
\setlength{\belowcaptionskip}{-0.1cm}
\begin{minipage}[b]{1.0\linewidth}
  \centering
  \centerline{\includegraphics[width=8.5cm]{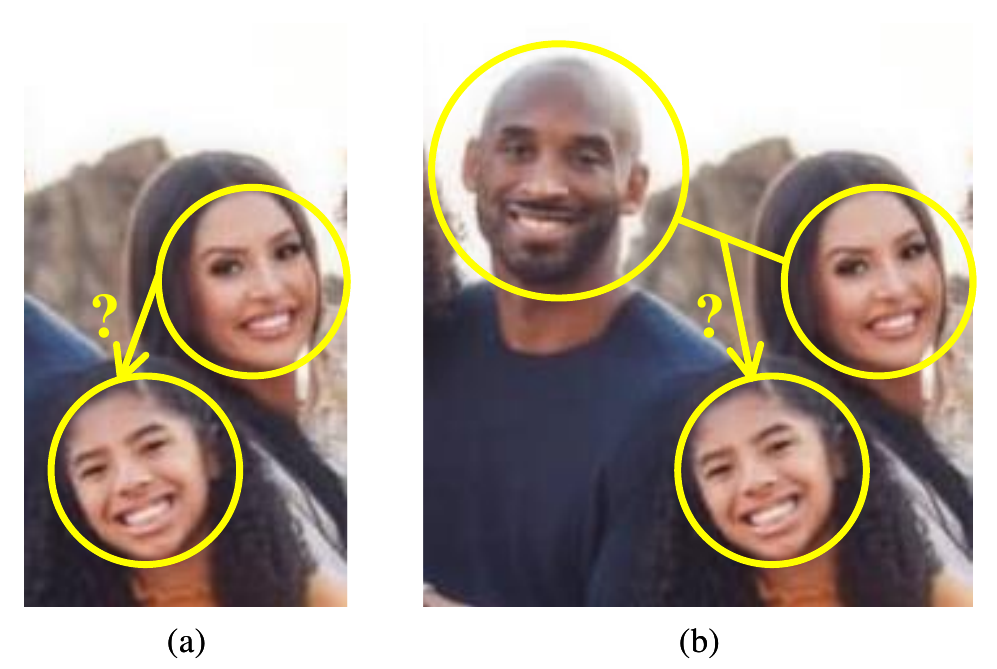}}
\end{minipage}
\caption{Schematic view of kinship verification. (a) 1-vs-1 kinship verification. (b) 2-vs-1 (tri-subject) kinship verification.}
\label{fig:fig1}
\end{figure}

The most basic kinship verification is 1-vs-1 verification, where two face images are given as inputs and the output is a decision whether the two persons are members of the same family, is shown in Fig. \ref{fig:fig1} (a). As a natural extension of 1-vs-1 verification, tri-subject verification is a special kinship verification problem of deciding whether a child is related to a pair of parents, which is essentially a 2-vs-1 verification, as shown in Fig. \ref{fig:fig1} (b). For example, when comparing two faces, one is a father and the other is a son, the prospective mother could be known if an image of the wife of the father is available. Thus, tri-subject verification aims to check the kinship between a pair of parents and a child.
 
Kinship verification has been proposed and extensively researched in the early years \cite{c3,c4,c5,c6,c24,c25,c26,c27,c28,c29,c30}. Fang et al. \cite{c3} first attempted kinship verification on parent-child face pairs by selecting several effective hand-crafted features to recognize kinship. Following this, some researchers \cite{c4, c5} recognized that a child’s face more closely resembles their parents at younger ages. They adopted transfer learning to 
narrow the appearance gap between the older faces and the 
younger faces. Lu et al. \cite{c6} used a metric learning method in euclidean space and its multi-view counterpart that learns a common distance metric for multiple feature types. Qin et al. \cite{c19} adopted a spatially voted method for feature selection and a relative symmetric bilinear model to simulate the similarity for tri-subject kinship verification.

Recently, deep learning \cite{c7} has been proved successful in a wide range of machine learning tasks. With the release of large-scale kinship datasets \cite{c8}, deep learning has achieved significant progress in several kinship recognition tasks \cite{c9,c10,c11,c13,c14,c15,c16,c31,c32,c33,c34,c35}. Dahan et al. \cite{c11} used two VGG-Face \cite{c12} models to extract face features, and then concatenated them and fed them to a fully-connected network for metric learning. Duan et al. \cite{c13} integrated multiple deep models to verify kinships. Li et al. \cite{c14} fine-tuned a pre-trained model on the FIW dataset using a soft triplet loss for backpropagation. Nandy et al. \cite{c15} used a siamese network for kinship recognition. Laiadi et al. \cite{c16} checked kinships by deep face descriptors and tensor features.

In this work, we present a deep siamese network to achieve kinship verification. The deep siamese network consists of two branches, which are used to extract the features of two input face images respectively. The extracted features are combined with different operations and then concatenated into a long vector. Next, the long vector is fed into a fully connected network to measure the relative similarity between two persons. Finally, we set a threshold parameter to divide the similarity score into 1 or 0, which indicates whether two persons are related or not. In order to benefit from different models, we introduce a jury system for multi-model fusion. In addition, we propose a deep triplet network for tri-subject kinship verification, which is constructed based on two deep siamese networks. The two deep siamese networks share a branch to extract the features of the child. Therefore, the deep triplet network has three branches for extracting the features of father, mother, and child respectively. The features are also fused by combining and concatenating, then they are fed into two fully connected networks to obtain the similarity scores of father-child (FC) and mother-child (MC). Finally, by a threshold parameter, the weighted sum of the two similarity scores is quantified as 1 or 0, which indicates whether a child is related to a pair of parents or not. The main contributions of this paper are summarized as follows.

\begin{figure}[t]
\setlength{\abovecaptionskip}{-0.0cm}
\setlength{\belowcaptionskip}{-0.24cm}
\begin{minipage}[b]{1.0\linewidth}
  \centering
  \centerline{\includegraphics[width=8.5cm]{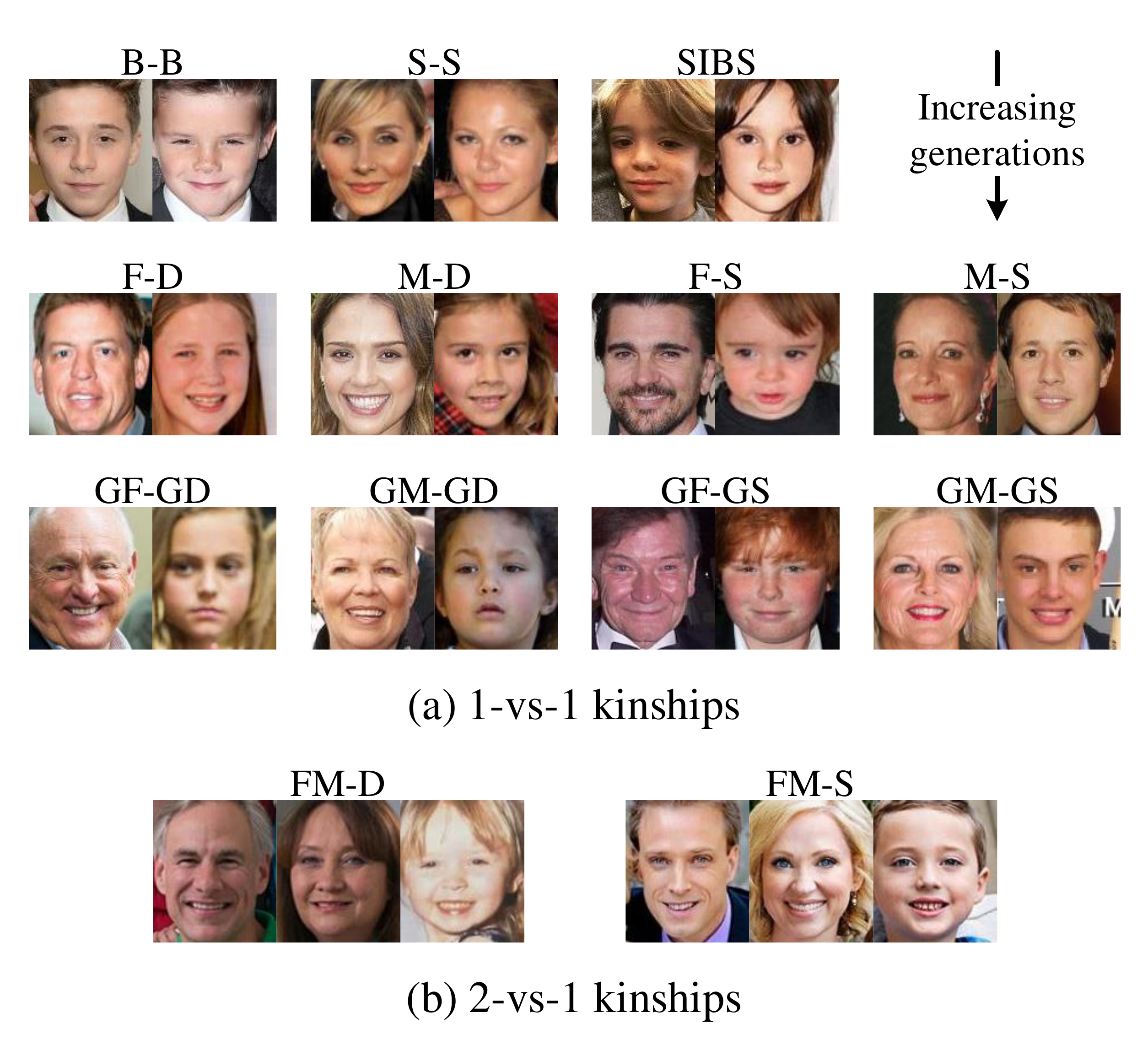}}
\end{minipage}
\caption{Some samples of FIW dataset, including relationship pairs and relationship triplets. (a) 11 types of 1-vs-1 kinships. (b) 2 types of 2-vs-1 (tri-subject) kinships.}
\label{fig:fig2}
\end{figure}



(1) A deep siamese network is proposed for bi-subject kinship verification. We employ multiple feature fusion operations to improve similarity metric learning in the network, instead of concatenating features directly \cite{c15}. Meanwhile, we adopt a jury system to benefit from multi-model fusion.

(2) By fusing two deep siamese networks, we present a deep triplet network for tri-subject kinship verification. Compared to \cite{c16}, several hyper-parameters are additionally used to flexibly improve the performance of the network. 

(3) Our team (ustc-nelslip) has already made remarkable achievements in the RFIW2020 Challenge, where we achieve 1st place in the Tri-Subject Verification (track II), and 3rd place in the Kinship Verification (track I).


\section{DATASET DESCRIPTION}

\begin{figure}[t]
\setlength{\abovecaptionskip}{0.0cm}
\setlength{\belowcaptionskip}{-0.0cm}
\begin{minipage}[b]{1.0\linewidth}
  \centering
  \centerline{\includegraphics[width=8.5cm]{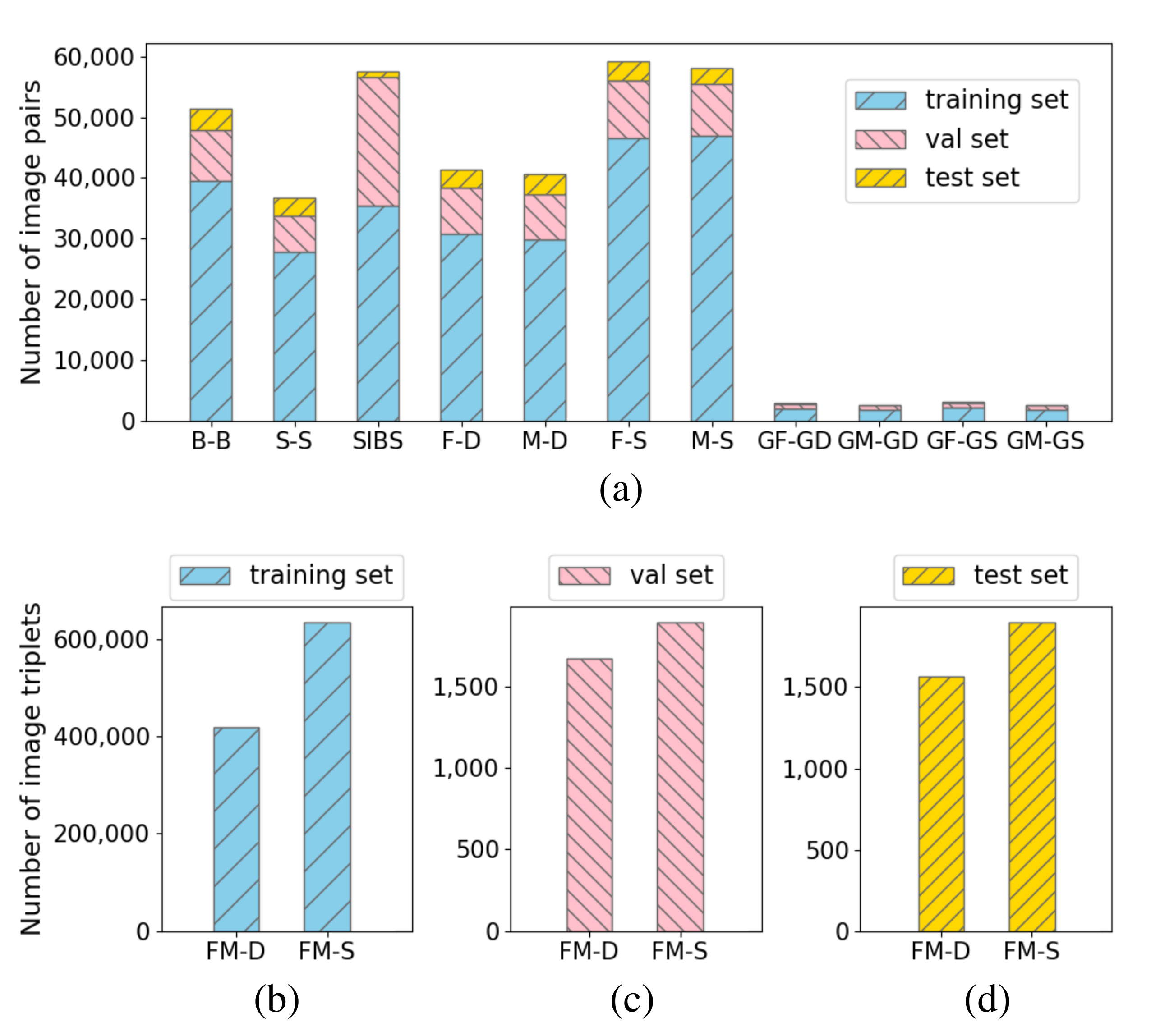}}
\end{minipage}
\caption{The data distribution of FIW dataset. (a) The number of image pairs in training set, val set and test set, including 11 types of 1-vs-1 kinships. From (b) to (d): The number of image triplets in training set, val set and test set, including 2 types of 2-vs-1 (tri-subject) kinships.}
\label{fig:fig3}
\end{figure}

Families In the Wild (FIW) \cite{c8} is the largest and most comprehensive image database for kinship recognition. Its motivation is to provide the resources needed for kinship recognition to transition from research-to-reality. With over 11,932 family photos of more than 1,000 families, FIW closely reflects the true data distribution of families worldwide. With the launch of Recognizing Families In the Wild (RFIW) Challenge \cite{c35}, FIW has received a lot of attention.

FIW dataset consists of 11 types of relationship pairs, i.e., brother-brother(B-B), sister-sister(S-S), brother-sister(SIBS), father-daughter(F-D), mother-daughter(M-D), father-son(F-S), mother-son(M-S), grandfather-granddaughter(GF-GD), grandmother-granddaughter(GM-GD), grandfather-grandson (GF-GS), grandmother-grandson(GM-GS), as shown in Fig. \ref{fig:fig2} (a). Thus, it contains relationships across three generations and is suitable for kinship verification research. These relationship pairs can be flexibly extended to relationship triplets (i.e., tri-subject kinships), like father-mother-daughter (FM-D) and father-mother-son (FM-S), as shown in Fig. \ref{fig:fig2} (b). Thus, it also applies to tri-subject verification research.

RFIW \cite{c35} is a large-scale kinship recognition challenge based on FIW, which supports multiple tracks, such as the kinship verification (track I) and tri-subject verification (track II). For track I, the goal is to determine whether a pair of faces are blood relatives. The data distribution of FIW in track I is illustrated in Fig. \ref{fig:fig3} (a), covering the number of image pairs in training set, val set and test set, including 11 types of relationship pairs. For track II, the main focus is to decide whether a child is related to a pair of parents. The data distribution of FIW in track II is shown in Fig. \ref{fig:fig3} (b) to (d), including  2 types of tri-subject kinships.



\begin{figure*}[htp]
\setlength{\abovecaptionskip}{0.0cm}
\setlength{\belowcaptionskip}{-0.0cm}
\begin{minipage}[b]{1.0\linewidth}
  \centering
  \centerline{\includegraphics[width=18.0cm]{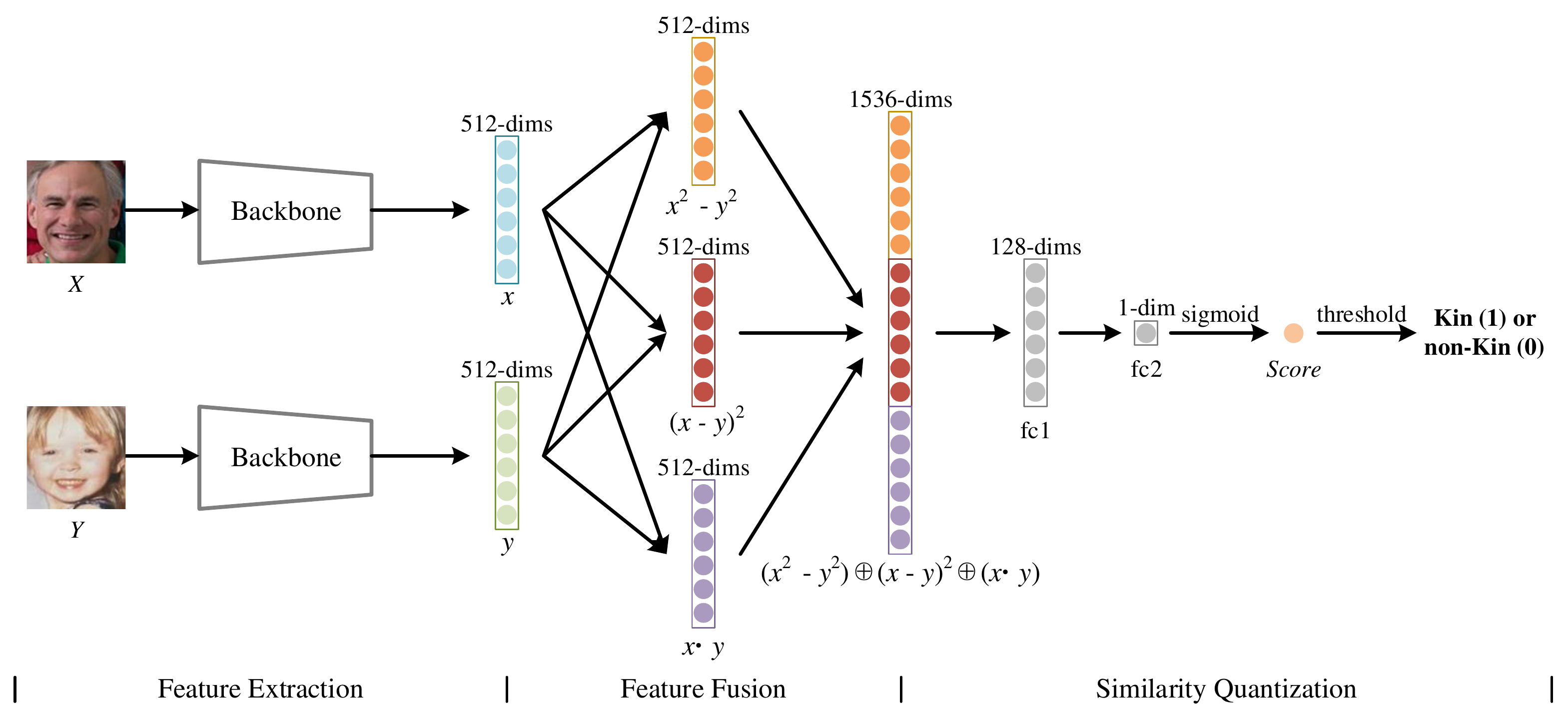}}
\end{minipage}
\caption{Overview of the proposed Deep Siamese Network. The deep siamese network consists of feature extraction, feature fusion and similarity quantization, where the schematic diagram of feature fusion takes ($x^2$-$y^2$)$\oplus$($x$-$y$)$^2$$\oplus$($x \cdot y$) as an example.}
\label{fig:fig4}
\end{figure*}

\section{PROPOSED METHOD}

Modern genetic studies clearly show that there is a high similarity of traits between family members, e.g., behavior and appearance. Therefore, it is possible to identify the kin relations between family members based on their face images. General kinship verification is a 1-vs-1 verification task, which is to check two persons are related by blood. Tri-subject verification, as a natural extension of 1-to-1 kinship verification, is a 2-vs-1 verification task, which is to verify whether a child is related to a pair of parents. In this work, we propose a Deep Siamese Network for 1-vs-1 (bi-subject) kinship verification and a Deep Triplet Network for 2-vs-1 (tri-subject) kinship verification.

\subsection{Deep Siamese Network}

The deep siamese network is built to verify the kinships between two individuals, where there are two branches with shared weights used to extract the features of two input face images respectively, as shown in Fig. \ref{fig:fig4}. In this work, we adopt ResNet50 \cite{c20} or SENet50 \cite{c21} as backbones for feature extraction, which are pre-trained on the VGGFace2 \cite{c17} dataset. The extracted features are then fused by combining and concatenating. Next, the fused features are fed into a fully-connected network to quantify the relative similarity score between two persons. By setting a threshold $t$, we can obtain the final predicted outcomes, Kin or non-Kin (i.e., 1 or 0 respectively). The final prediction is defined as
\begin{equation}
\setlength{\abovedisplayskip}{0.2cm}
\setlength{\belowdisplayskip}{0.2cm}
Prediction=\left\{
\begin{aligned}
1 & , & if\ Score>t, \\
0 & , & otherwise.
\end{aligned}
\right.
\label{eq:eq1}
\end{equation}

The proposed deep siamese network can verify 11 types of 1-vs-1 kinships. The verification accuracy, as a common metric, is employed to evaluate the performance of relevant methods, which can be formulated as

\begin{equation}
\setlength{\abovedisplayskip}{0.2cm}
\setlength{\belowdisplayskip}{0.2cm}
{Acc_j} = \frac {Correct\ Predictions\ for\ j^{th}\ type} {Total\ of\ pairs\ for\ j^{th}\ type}
\end{equation}
where $j^{th} \in$ \{all 11 relationship types\}. The overall accuracy is calculated as a weighted sum (i.e., weight by the pair count to determine the average accuracy).

\subsection{Feature Fusion}

Feature fusion is helpful to improve the nonlinear ability of the deep siamese network. Furthermore, the essence of feature fusion is to encode two input face features abstractly, which is beneficial for the fully-connected network to learn better similarity metrics. Suppose that $X$ and $Y$ denote the two input face images, $x$ and $y$ denote the extracted features of two faces, $\oplus$ represents the feature concatenation operation. In this work, there are 5  different types of feature fusion, i.e., $x \oplus y$, ($x$+$y$)$\oplus$($x$-$y$), ($x$+$y$)$\oplus$($x$-$y$)$\oplus$($x\cdot y$), ($x^2$-$y^2$)$\oplus$($x$-$y$)$^2$, ($x^2$-$y^2$)$\oplus$($x$-$y$)$^2$$\oplus$($x\cdot y$). Fig. \ref{fig:fig4} shows the details of feature fusion, taking ($x^2$-$y^2$)$\oplus$($x$-$y$)$^2$$\oplus$($x\cdot y$) as an example.

The extracted face features $x$ and $y$ undergo a variety of operations such as element-wise multiplication (e.g., $x^2$, $y^2$, $x\cdot y$), element-wise addition ($x$+$y$), and element-wise subtraction ($x$-$y$, $x^2$-$y^2$). After that, they are fused by combining and concatenating. Next, the fused features are fed into a fully-connected network. The fully-connected network is built with two fully-connected layers and a sigmoid activation layer, where the first fully-connected layer (fc1) is 128 dimensions and the second (fc2) is only 1 dimension. The output of fc2 is activated by a sigmoid function, obtaining the similarity score for deciding whether the two persons are related.

\begin{figure*}[t]
\setlength{\abovecaptionskip}{0.0cm}
\setlength{\belowcaptionskip}{-0.0cm}
\begin{minipage}[b]{1.0\linewidth}
  \centering
  \centerline{\includegraphics[width=18.0cm]{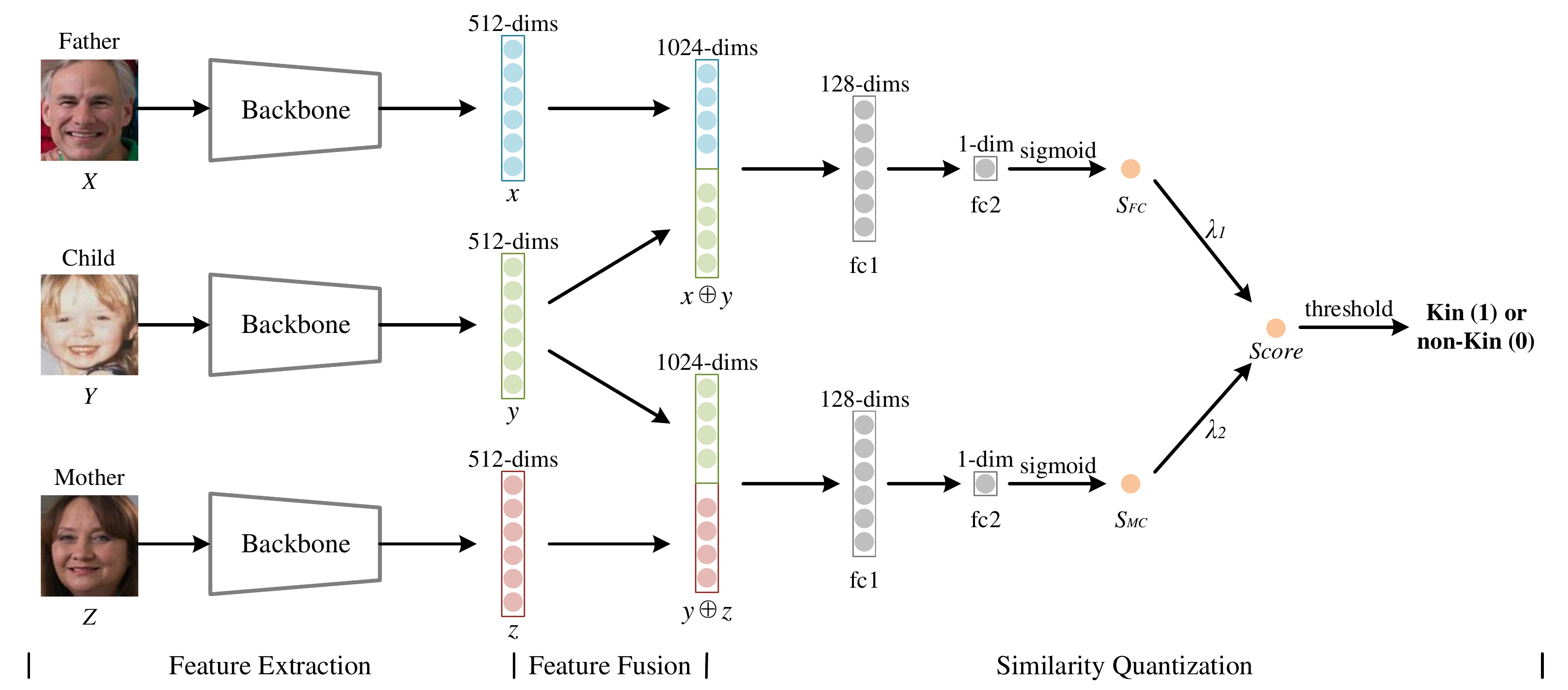}}
\end{minipage}
\caption{Overview of the proposed Deep Triplet Network. The deep triplet network consists of feature extraction, feature fusion and similarity quantization, where the schematic diagram of feature fusion takes $x\oplus y$ and $y\oplus z$ as examples.}
\label{fig:fig5}
\end{figure*}

\subsection{Loss function}

In this work, two different loss functions are tried out, i.e., binary-cross entropy (\emph{BCE}) loss and focal loss \cite{c18}.

\textbf{\emph{BCE loss.}} The binary cross entropy loss is given as
\begin{equation}
\setlength{\abovedisplayskip}{0.2cm}
\setlength{\belowdisplayskip}{0.2cm}
{\mathcal{L}_{bce}} = -{y}\log({p}) - (1-{y})\log({1-p}),
\end{equation}
where $y$ is the target label that is either 0 or 1, and $p$ is the predicted value by a sigmoid activation function.

\textbf{\emph{Focal loss.}} The focal loss is helpful to alleviate the significant imbalance of the proportion of positive and negative samples, which can be formulated as
\begin{equation}
\setlength{\abovedisplayskip}{0.2cm}
\setlength{\belowdisplayskip}{0.2cm}
{\mathcal{L}_{focal}} = -{a_t}({1-p_t})^{\gamma}\log({p_t}),
\end{equation}

\begin{equation}
\setlength{\abovedisplayskip}{0.2cm}
\setlength{\belowdisplayskip}{0.2cm}
a_{t} = \left \{
\begin{aligned}
a ,&\ if\ y=1 \\
1 - a ,&\ otherwise
\end{aligned}
\right.
\end{equation}

\begin{equation}
\setlength{\abovedisplayskip}{0.2cm}
\setlength{\belowdisplayskip}{0.2cm}
p_{t} = \left \{
\begin{aligned}
p ,&\ if\ y=1 \\
1 - p ,&\ otherwise
\end{aligned}
\right.
\end{equation}
where $\gamma$ is a focusing parameter, $a$ is a weight that controls the contribution of positive and negative samples to the total loss, and $p$ is the output of the sigmoid activation function. When $a=0.5$, the focal loss is equivalent to \emph{BCE} loss. 

\subsection{Jury System}

In order to benefit from models with different configurations such as backbones, losses, and feature fusions, we adopt a jury system for multi-model fusion. Specifically, we choose the best single model as the major model and several excellent models as the auxiliary models. Moreover, we set 3 thresholds, i.e., a low value, a median value, and a high value. In this work, the low value is $0.1$, the median value is $0.3$, and the high value is $0.5$. These values are used to divide into different intervals, corresponding to different calculation methods of the predictions. The similarity score of the major model is compared with the low value and the high value to obtain the binarized predictions. The average of the similarity scores of multiple models is compared with the median value to obtain the predicted results. The following pseudocode Algorithm \ref{alg:alg1} shows the procedures of the jury system.

\begin{algorithm}[t]
 \caption{The Pipline of Jury System.}
\label{alg:alg1}
 \LinesNumbered 
 \KwIn{a pair of input face images: $x$ and $y$; a major model: $M_1$; 3 auxiliary models: $M_2$, $M_3$ and $M_4$; 3 thresholds: $t_{low}$, $t_{median}$ and $t_{high}$.}
 \KwOut{a Boolean value: $Prediction$ = 1 or 0 (Kin or non-Kin respectively).}
 \If{ $M_1(x, y)<t_{low}$ }
 {
  $Prediction=0$\;
 }
 \ElseIf{ $M_1(x, y)>t_{high}$ }
 {
  $Prediction=1$\;
 }
 \ElseIf{ mean($M_1(x, y) + M_2(x, y) + M_3(x, y) + M_4(x, y))$ 
				 \ \ \ \ $<t_{median}$ }
 {
  $Prediction=0$\;
 }
 \ElseIf{ mean($M_1(x, y) + M_2(x, y) + M_3(x, y) + M_4(x, y))$ 
				 \ \ \ \ $>t_{median}$ }
 {
  $Prediction=1$\;
 }
\end{algorithm}

\subsection{Deep Triplet Network}

Tri-subject verification is a special kind of kinship verification task, because when you know one parent, the other parent is likely accessible. Therefore, tri-subject verification is to determine whether a child is related to a pair of parents. The deep triplet network is built based on two deep siamese networks for tri-subject kinship verification, which can learn two kinship pairs, i.e., father-child (FC) and mother-child (MC). As shown in Fig. \ref{fig:fig5}, the two siamese networks share the same branch to extract the features from a child image. Therefore, there are three branches without sharing weights employed to extract the features of father, mother, and child respectively. We also introduce ResNet50 or SENet50 as a backbone for feature extraction, which are pre-trained on the VGGFace2 dataset. The extracted child features are fused with the extracted features of father and mother respectively. Then the fused features are fed into fully-connected networks for obtaining the similarity scores, i.e., $S_{FC}$ and $S_{MC}$. The operations of feature fusion and similarity quantization are consistent with that of the deep siamese network. By weighting and summing the two similarity scores, the final parent-child similarity score is obtained as
\begin{equation}
\setlength{\abovedisplayskip}{0.2cm}
\setlength{\belowdisplayskip}{0.2cm}
Score = {\lambda}{_1}S_{FC} + {\lambda}{_2}S_{MC},
\end{equation}
where ${\lambda}{_1}$ and ${\lambda}{_1}$ are the hyper-parameters for balancing the relative importance between $S_{FC}$ and $S_{MC}$. In addition, a threshold $t$ is used to output the final predicted results, i.e., Kin or non-Kin, formulated as Eq. \ref{eq:eq1}.

\section{EXPERIMENTS}

\begin{table}[]
\small
\setlength{\abovecaptionskip}{0.0cm}
\setlength{\belowcaptionskip}{0.0cm}
\renewcommand\arraystretch{1.2}
\linespread{0.8}
\centering
\caption{Number of image pairs or triplets on FIW dataset.}
\setlength{\tabcolsep}{5.1mm}{
\begin{tabular}{cccc}
\toprule
\multirow{2}{*}{Kinships} & \multicolumn{3}{c}{Number of image pairs or triplets} \\ \cmidrule{2-4} 
                          & training set         & val set         & test set       \\ \midrule 
B-B                       & 39,608               & 8,340           & 3,459          \\ 
S-S                       & 27,844               & 5,982           & 2,956          \\
SIBS                      & 35,337               & 21,204          & 967            \\
F-D                       & 30,746               & 7,575           & 3,019          \\
M-D                       & 29,778               & 7,587           & 3,273          \\
F-S                       & 46,583               & 9,399           & 3,184          \\
M-S                       & 46,969               & 8,441           & 2,660          \\
GF-GD                     & 2,003                & 762             & 121            \\
GM-GD                     & 1,741                & 714             & 71             \\
GF-GS                     & 2,097                & 879             & 96             \\
GM-GS                     & 1,834                & 701             & 84             \\
FM-D                      & 419,892              & 1,673           & 1,569          \\
FM-S                      & 635,492              & 1,895           & 1,901          \\ \bottomrule
\end{tabular}}
\label{tab:tab1}
\end{table}

A workstation with Intel i7-7700K 4.2G CPU, 64G memory and NVIDIA GTX2080 8G GPU is used for the experiments. We evaluate our method on the FIW dataset, and the essential ablation studies are elaborately designed, as well as quantitative evaluations with other contestants. 

For implementation details, all face images are resized to 224$\times$224 pixels as the inputs. The numbers of image pairs or triplets for the training set, val set, and test set are shown in Tabel \ref{tab:tab1}. An equal number of negative samples are constructed for training. For optimization, we choose the Adam optimizer \cite{c22}, where the initial learning rate is set to $0.001$ and other settings are default. The model is trained with a mini-batch size of 32 and over 60 epochs, which is based on Keras using TensorFlow backend. In addition, we evaluate our method on the val set of FIW dataset and compare it with other contestants on the test set of FIW dataset.

\begin{table}[t]
\small
\setlength{\abovecaptionskip}{0.0cm}
\setlength{\belowcaptionskip}{0.0cm}
\renewcommand\arraystretch{1.3}
\linespread{0.8}
\centering
\caption{Ablation Studies of Deep Siamese Network.}
\setlength{\tabcolsep}{1.8mm}{
\begin{tabular}{cccc}
\toprule
Backbone                   &  Loss                       & Feature Fusion                      & Accuracy       \\  \midrule
\multirow{4}{*}{ResNet50}  & \multirow{2}{*}{BCE loss}   & ($x$+$y$)$\oplus$($x$-$y$)$\oplus$($x\cdot y$)
& 0.725          \\  
                           &                             & ($x^2$-$y^2$)$\oplus$($x$-$y$)$^2$$\oplus$($x\cdot y$)  & {\color{red}0.738}          \\ \cmidrule{2-4} 
                           & \multirow{2}{*}{Focal loss} & ($x$+$y$)$\oplus$($x$-$y$)$\oplus$($x\cdot y$)
& 0.721          \\   
                           &                             & ($x^2$-$y^2$)$\oplus$($x$-$y$)$^2$$\oplus$($x\cdot y$)  & {\color{blue}0.735}          \\ \cmidrule{1-4}
\multirow{4}{*}{SENet50}   & \multirow{2}{*}{BCE loss}   & ($x$+$y$)$\oplus$($x$-$y$)$\oplus$($x\cdot y$)
& 0.712          \\   
                           &                             & ($x^2$-$y^2$)$\oplus$($x$-$y$)$^2$$\oplus$($x\cdot y$)  & 0.733          \\ \cmidrule{2-4} 
                           & \multirow{2}{*}{Focal loss} & ($x$+$y$)$\oplus$($x$-$y$)$\oplus$($x\cdot y$)
& 0.698          \\  
                           &                             & ($x^2$-$y^2$)$\oplus$($x$-$y$)$^2$$\oplus$($x\cdot y$)  & 0.719          \\ \bottomrule
\end{tabular}}
\label{tab:tab2}
\end{table}

\begin{table}[t]
\small
\setlength{\abovecaptionskip}{0.0cm}
\setlength{\belowcaptionskip}{0.0cm}
\renewcommand\arraystretch{1.2}
\linespread{0.8}
\centering
\caption{The Effect of Jury System for model fusion.}
\setlength{\tabcolsep}{2.0mm}{
\begin{tabular}{ccc}
\toprule
\multirow{2}{*}{Kinships} & \multicolumn{2}{c}{Using jury system or not ($w/$ js or $w/o$ js)} \\ \cmidrule{2-3} 
                          & best single model ($w/o$ js)  & multi-model ($w/$ js)     \\ \midrule
B-B                       & 0.729               & 0.751               \\ 
S-S                       & 0.732               & 0.744              \\
SIBS                      & 0.714               & 0.721              \\
F-D                       & 0.741               & 0.755               \\
M-D                       & 0.725               & 0.747              \\
F-S                       & 0.783               & 0.818              \\
M-S                       & 0.740               & 0.752              \\
GF-GD                     & 0.682               & 0.786              \\
GM-GD                     & 0.651               & 0.758              \\
GF-GS                     & 0.673               & 0.691              \\
GM-GS                     & 0.648               & 0.670              \\ 
Average                   & 0.738               & 0.759              \\ \bottomrule
\end{tabular}}
\label{tab:tab3}
\end{table}

\subsection{Ablation Studies of Deep Siamese Network}

In this work, we implement the ablation studies of the deep siamese network based on 2 backbones (i.e., ResNet50 and SENet50), 2 loss functions (i.e., binary-cross entropy loss and focal loss) and 2 types of feature fusions (i.e., ($x$+$y$)$\oplus$($x$-$y$)$\oplus$($x\cdot y$) and ($x^2$-$y^2$)$\oplus$($x$-$y$)$^2$$\oplus$($x\cdot y$)). As shown in Table \ref{tab:tab2}, a total of 8 combinations are evaluated on the val set of FIW dataset, where the best result is marked red and the second-best is marked blue. When the backbone is ResNet50, the loss function is BCE loss, and the feature fusion operation is ($x^2$-$y^2$)$\oplus$($x$-$y$)$^2$$\oplus$($x\cdot y$), we obtain the highest verification accuracy, which is up to $0.738$. In addition, experimental results show that BCE loss performs better than the focal loss in most cases, and ResNet50 is superior to SENet50 as a backbone in the deep siamese network.

\begin{table*}[t]
\small
\setlength{\abovecaptionskip}{0.0cm}
\setlength{\belowcaptionskip}{0.0cm}
\renewcommand\arraystretch{1.4}
\linespread{0.8}
\centering
\caption{Leaderboard of RFIW Kinship Verification (Top 10 contestants and baseline).}
\setlength{\tabcolsep}{1.2mm}{
\begin{tabular}{ccccccccccccc}
\toprule
\multirow{2}{*}{user}   & \multicolumn{12}{c}{Verification Accuracy}  \\  \cmidrule{2-13}
		      & BB         & SS        & SIBS           & FD       & MD     	& FS       & MS       & GFGD     & GMGD     & GFGS     & GMGS     & Average  \\  \midrule
vuvko        & {\color{red}0.80}(1)  & {\color{red}0.80}(1)   & {\color{red}0.77}(1)    & 0.75(4)  & {\color{red}0.78}(1)	& 0.81(4)  & 0.74(6)  & {\color{blue}0.78}(2)  & {\color{red}0.76}(1)  & 0.69(5)  & 0.60(8)  & {\color{red}0.78}(1)  \\
DeepBlueAI   & {\color{blue}0.77}(2)  & {\color{blue}0.77}(2)  & 0.75(3)   & 0.74(5)  & 0.75(4)	& 0.81(5)  & 0.74(7)  & 0.72(5)  & 0.67(7)  & 0.73(3)  & {\color{red}0.68}(1)  & {\color{blue}0.76}(2)  \\
ustc-nelslip & 0.75(3)  & 0.74(4)  & 0.72(6)  & {\color{blue}0.76}(2)  & 0.75(5)	& {\color{red}0.82}(1)  & 0.75(3)  & {\color{red}0.79}(1)  & {\color{red}0.76}(1)  & 0.69(6)  & {\color{blue}0.67}(2)  & 0.76(3)  \\
haoxl        & 0.75(5)  & 0.74(6)  & 0.71(8)  & {\color{blue}0.76}(2)  & 0.75(5)	& 0.81(3)  & 0.75(3)  & 0.73(4)  & 0.63(11) & 0.65(10) & 0.64(5)  & 0.76(4)  \\
lemoner20    & 0.75(4)  & 0.74(3)  & 0.72(5)  & 0.75(3)  & 0.74(6)	& 0.81(6)  & 0.75(4)  & 0.72(6)  & 0.62(13) & 0.67(8)  & 0.65(3)  & 0.75(5)  \\
Early        & 0.75(6)  & 0.74(5)  & 0.73(4)  & 0.73(6)  & 0.72(8)	& 0.79(8)  & 0.74(5)  & 0.66(15) & 0.52(16) & 0.69(7)  & 0.65(4)  & 0.74(6)  \\
stefhoer     & 0.66(13) & 0.65(13) & {\color{blue}0.76}(2)   & {\color{red}0.77}(1)  & {\color{blue}0.77}(2)	& 0.80(7)  & {\color{red}0.78}(1)  & 0.70(10) & 0.64(10) & {\color{blue}0.73}(2)  & 0.60(7)  & 0.74(7)  \\
bestone      & 0.69(11) & 0.67(12) & 0.62(13) & 0.75(3)  & 0.75(3)	& {\color{blue}0.81}(2)  & {\color{blue}0.75}(2)  & 0.73(3)  & 0.65(9)  & 0.69(5)  & 0.62(6)  & 0.73(8)  \\
danbo3004    & 0.71(8)  & 0.72(10) & 0.71(9)  & 0.72(7)  & 0.72(10)	& 0.78(9)  & 0.72(8)  & 0.71(8)  & 0.53(15) & 0.70(4)  & 0.56(11) & 0.73(9)  \\
ten\_elven   & 0.72(7)  & 0.73(7)  & 0.71(7)  & 0.70(8)  & 0.70(11)	& 0.77(10) & 0.71(9)  & 0.70(11) & 0.63(11) & {\color{red}0.75}(1)  & {\color{blue}0.67}(2)  & 0.72(10) \\
baseline     & 0.71(-)  & 0.73(-)  & 0.66(-)  & 0.61(-)  & 0.69(-) & 0.66(-)  & 0.62(-)  & 0.68(-)  & 0.64(-)  & 0.57(-)  & 0.50(-)  & 0.64(-)   \\
\bottomrule
\end{tabular}}
\label{tab:tab4}
\end{table*}

\begin{figure}[t]
\setlength{\abovecaptionskip}{0.0cm}
\setlength{\belowcaptionskip}{-0.1cm}
\begin{minipage}[b]{1.0\linewidth}
  \centering
  \centerline{\includegraphics[width=8.6cm]{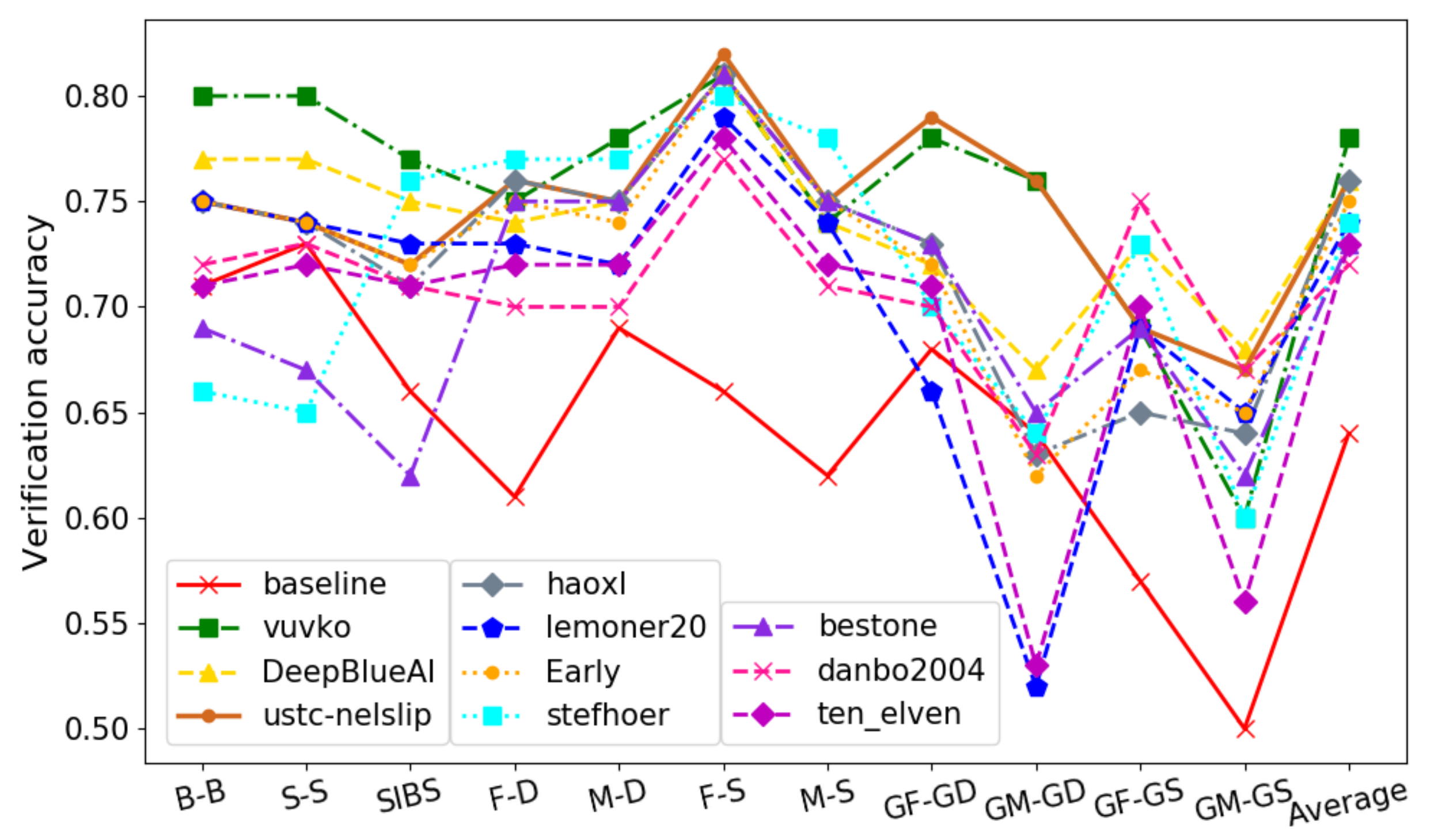}}
\end{minipage}
\caption{Verification accuracy line chart for 11 types of kinship pairs, including top 10 contestants in the leaderboard.}
\label{fig:fig6}
\end{figure}

\begin{figure}[t]
\setlength{\abovecaptionskip}{0.0cm}
\setlength{\belowcaptionskip}{-0.0cm}
\begin{minipage}[b]{1.0\linewidth}
  \centering
  \centerline{\includegraphics[width=8.6cm]{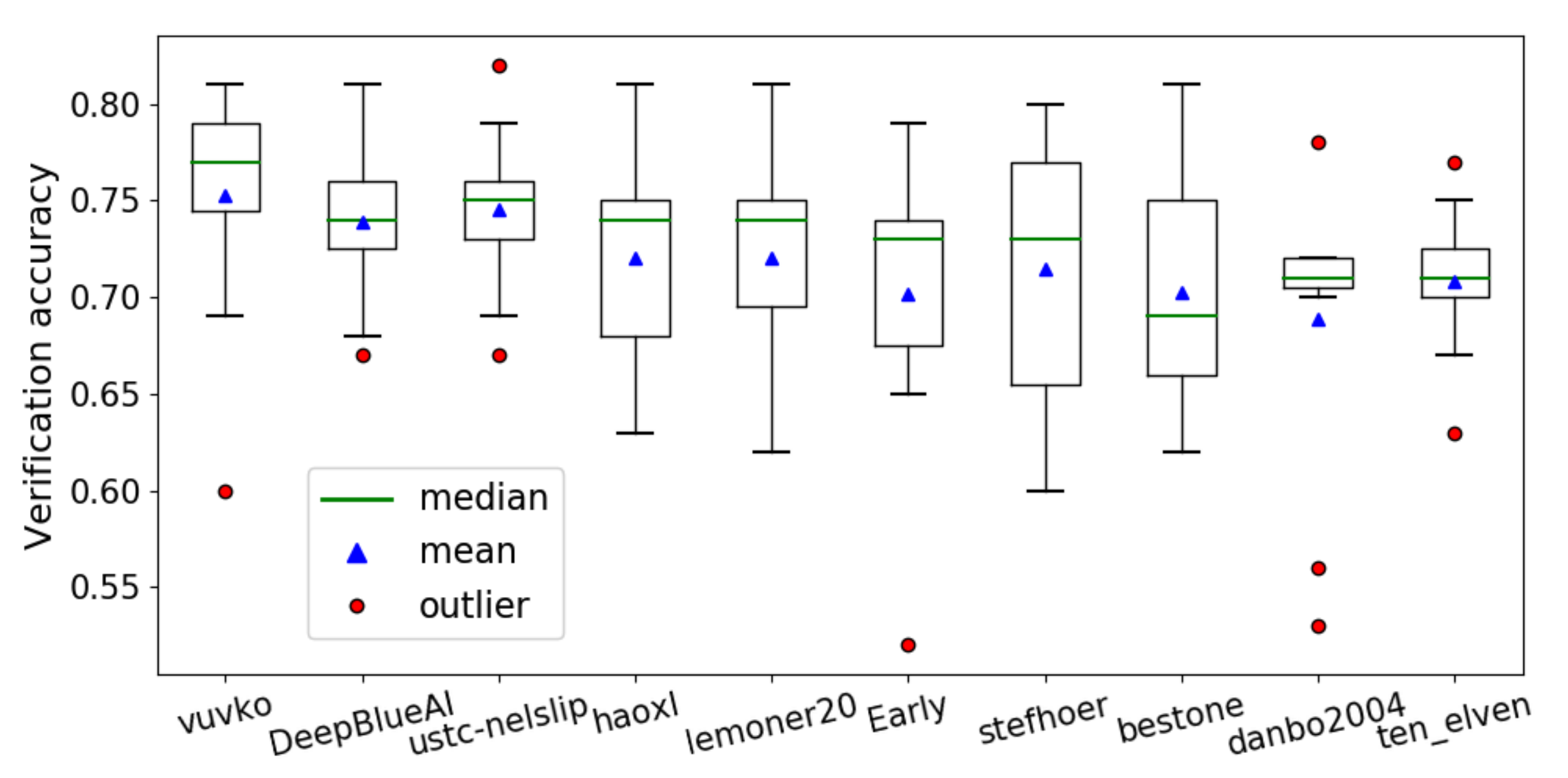}}
\end{minipage}
\caption{Verification accuracy boxplot for 11 types of kinship pairs, including top 10 contestants in the leaderboard.}
\label{fig:fig7}
\end{figure}

\subsection{The Effect of Jury System}

In order to benefit from multiple models, we introduce a jury system for fusing different models. As shown in Table \ref{tab:tab3}, we evaluate the best single model and the fused multi-model on the test set of FIW dataset. Experimental results show that the verification accuracy of the best single model is $0.738$, and the multi-model is $0.759$. Thus, the performance of the multi-model fused by the jury system is superior to the best single model in all kinship pairs.

\begin{table}[t]
\small
\setlength{\abovecaptionskip}{0.0cm}
\setlength{\belowcaptionskip}{-0.0cm}
\renewcommand\arraystretch{1.2}
\linespread{0.8}
\centering
\caption{The Effect of Hyper-parameter ${\lambda}{_1}$ and ${\lambda}{_2}$.}
\setlength{\tabcolsep}{4.8mm}{
\begin{tabular}{ccccc}
\toprule
\multirow{2}{*}{${\lambda}{_1}$} & \multirow{2}{*}{${\lambda}{_2}$} & \multicolumn{3}{c}{Verification Accuracy} \\ \cmidrule{3-5} 
                    &               & FM-D           & FM-S           & Average        \\ \midrule
0.3                 & 0.7           & 0.653          & 0.642          & 0.648          \\ 
0.4                 & 0.6           & 0.682          & 0.669          & 0.676          \\ 
0.5                 & 0.5           & \textbf{0.712} & 0.719          & \textbf{0.716} \\ 
0.6                 & 0.4           & 0.687          & \textbf{0.723} & 0.705          \\ 
0.7                 & 0.3           & 0.647          & 0.663          & 0.655          \\  \bottomrule
\end{tabular}}
\label{tab:tab6}
\end{table}

\begin{table}[t]
\small
\setlength{\abovecaptionskip}{0.0cm}
\setlength{\belowcaptionskip}{0.0cm}
\renewcommand\arraystretch{1.2}
\linespread{0.8}
\centering
\caption{The Effect of Hyper-parameter $t$.}
\setlength{\tabcolsep}{6.5mm}{
\begin{tabular}{cccc}
\toprule
\multirow{2}{*}{$t$}      & \multicolumn{3}{c}{Verification Accuracy}            \\ \cmidrule{2-4} 
                          & FM-D           & FM-S           & Average             \\ \midrule
0.2                       & 0.652          & 0.634          & 0.643               \\ 
0.3                       & \textbf{0.714} & 0.702          & 0.708               \\ 
0.4                       & 0.712          & \textbf{0.719} & \textbf{0.716}      \\ 
0.5                       & 0.638          & 0.646          & 0.642               \\ 
0.6                       & 0.561          & 0.565          & 0.563               \\ \bottomrule
\end{tabular}}
\label{tab:tab7}
\end{table}

\subsection{Quantitative Evaluations of Kinship Verification}

The leaderboard of RFIW2020 Challenge in Kinship Verification (track I) is illustrated in Table  \ref{tab:tab4}, where we only show the top 10 contestants and the baseline. For our method, the average verification accuracy of all types of kinship pairs is up to $0.76$, achieving the 3rd place. The line chart of verification accuracy is drawn in Fig. \ref{fig:fig6}, where the accuracy of F-S is the highest of all kinship pairs, and that of GM-GS and GM-GD are unstable. As shown in Fig. \ref{fig:fig7}, the mean, median and standard deviations of the boxplot reflect the comprehensive performance of each method for all kinships.

\begin{table*}[t]
\small
\setlength{\abovecaptionskip}{0.0cm}
\setlength{\belowcaptionskip}{0.0cm}
\renewcommand\arraystretch{1.25}
\linespread{0.8}
\centering
\caption{Ablation Studies of Deep Triplet Network.}
\setlength{\tabcolsep}{7.8mm}{
\begin{tabular}{cccccc}
\toprule
\multirow{2}{*}{Backbone} & \multirow{2}{*}{Loss}     & \multirow{2}{*}{Feature Fusion}    & \multicolumn{3}{c}{Verification Accuracy}     \\ \cmidrule{4-6} 
                           &                             &                                           & FM-D          & FM-S          & Average       \\ \midrule
\multirow{10}{*}{ResNet50} & \multirow{5}{*}{BCE loss}   & $x \oplus y$
& 0.672         & 0.676         & 0.674          \\  
                           &                             & ($x$+$y$)$\oplus$($x$-$y$)                        & 0.688         & 0.685         & 0.687          \\ 
                           &                             & ($x$+$y$)$\oplus$($x$-$y$)$\oplus$($x\cdot y$)
& 0.699         & 0.705         & 0.702          \\ 
                           &                             & ($x^2$-$y^2$)$\oplus$($x$-$y$)$^2$        & 0.704         & 0.708         & 0.706          \\ 
                           &                             & ($x^2$-$y^2$)$\oplus$($x$-$y$)$^2$$\oplus$($x\cdot y$) &{\color{red}0.712} &{\color{red}0.719} &{\color{red}0.716}  \\  \cmidrule{2-6}
                           & \multirow{5}{*}{Focal loss} & $x \oplus y$                                & 0.657         & 0.668         & 0.663          \\  
                           &                             & ($x$+$y$)$\oplus$($x$-$y$)                        & 0.670         & 0.676         & 0.673          \\  
                           &                             & ($x$+$y$)$\oplus$($x$-$y$)$\oplus$($x\cdot y$)             & 0.685         & 0.692         & 0.689          \\ 
                           &                             & ($x^2$-$y^2$)$\oplus$($x$-$y$)$^2$        & 0.691         & 0.704         & 0.698          \\  
                           &                             & ($x^2$-$y^2$)$\oplus$($x$-$y$)$^2$$\oplus$($x\cdot y$)  & {\color{blue}0.708}         & {\color{blue}0.715}         & {\color{blue}0.712}          \\ \cmidrule{1-6}
\multirow{10}{*}{SENet50}  & \multirow{5}{*}{BCE loss}   & $x \oplus y$                                & 0.634         & 0.642         & 0.638          \\ 
                           &                             & ($x$+$y$)$\oplus$($x$-$y$)                        & 0.645         & 0.662         & 0.654          \\  
                           &                             & ($x$+$y$)$\oplus$($x$-$y$)$\oplus$($x\cdot y$)              & 0.667         & 0.671         & 0.669          \\  
                           &                             & ($x^2$-$y^2$)$\oplus$($x$-$y$)$^2$        & 0.673         & 0.678         & 0.676          \\  
                           &                             & ($x^2$-$y^2$)$\oplus$($x$-$y$)$^2$$\oplus$($x\cdot y$)  & 0.689         & 0.693         & 0.691          \\ \cmidrule{2-6} 
                           & \multirow{5}{*}{Focal loss} & $x \oplus y$
& 0.626         & 0.634         & 0.630          \\ 
                           &                             & ($x$+$y$)$\oplus$($x$-$y$)
& 0.641         & 0.648         & 0.645          \\  
                           &                             & ($x$+$y$)$\oplus$($x$-$y$)$\oplus$($x\cdot y$)              & 0.656         & 0.661         & 0.659          \\  
                           &                             & ($x^2$-$y^2$)$\oplus$($x$-$y$)$^2$        & 0.670         & 0.673         & 0.672          \\  
                           &                             & ($x^2$-$y^2$)$\oplus$($x$-$y$)$^2$$\oplus$($x\cdot y$)  & 0.678         & 0.683         & 0.681          \\ \bottomrule
\end{tabular}}
\label{tab:tab5}
\end{table*}

\subsection{The Effect of Hyper-parameters}

In the deep triplet network, several hyper-parameters are set to improve the final outcomes. As shown in Table \ref{tab:tab6}, changing ${\lambda}{_1}$ and ${\lambda}{_2}$ has the effects on kinship verification accuracy. With the increase of ${\lambda}{_1}$, the accuracy of FM-S first increases and then decreases. One possible explanation is that non-relatives with the same gender are sometimes more similar than relatives with the opposite gender. As shown in Table \ref{tab:tab7}, different $t$ have different effects on the accuracy of FM-D and FM-S. When $t=0.3$, the accuracy of FM-D is best, while that of FM-S is best when $t=0.4$. The best average accuracy on the test set is up to $0.79$ when ${\lambda}{_1}=0.5$, ${\lambda}{_2}=0.5$, $t=0.3$ in FM-D and $t=0.4$ in FM-S.

\begin{table}[t]
\small
\setlength{\abovecaptionskip}{0.0cm}
\setlength{\belowcaptionskip}{0.0cm}
\renewcommand\arraystretch{1.4}
\linespread{0.8}
\centering
\caption{Leaderboard of RFIW Tri-Subject Verification.}
\setlength{\tabcolsep}{4.8mm}{
\begin{tabular}{cccc}
\toprule
\multirow{2}{*}{user} & \multicolumn{3}{c}{Verification Accuracy} \\ \cmidrule{2-4} 
                      & FM-D         & FM-S         & Average      \\ \midrule
ustc-nelslip          & {\color{red}0.78} (1)     & {\color{red}0.80} (1)     & {\color{red}0.79} (1)     \\ 
lemoner20             & 0.76 (4)     & {\color{blue}0.80} (2)     & {\color{blue}0.78} (2)     \\ 
DeepBlueAI            & {\color{blue}0.76} (2)     & 0.77 (3)     & 0.77 (3)     \\ 
Early                 & 0.76 (3)     & 0.77 (4)     & 0.77 (4)     \\ 
stefhoer              & 0.72 (5)     & 0.74 (6)     & 0.73 (5)     \\ 
Ferryman              & 0.70 (6)     & 0.74 (5)     & 0.72 (6)     \\ 
will\_go              & 0.66 (7)     & 0.70 (7)     & 0.68 (7)     \\ 
baseline              & 0.68 (-)     & 0.68 (-)     & 0.68 (-)     \\
\bottomrule
\end{tabular}}
\label{tab:tab8}
\end{table}

\subsection{Ablation Studies of Deep Triplet Network}

For the deep triplet network, we do ablation studies based on 2 backbones (i.e., ResNet50 \cite{c20} and SENet50 \cite{c21}), 2 loss functions (i.e., binary-cross entropy loss and focal loss) and 5 types of feature fusion operations (i.e., $x \oplus y$, ($x$+$y$)$\oplus$($x$-$y$), ($x$+$y$)$\oplus$($x$-$y$)$\oplus$($x\cdot y$), ($x^2$-$y^2$)$\oplus$($x$-$y$)$^2$ and ($x^2$-$y^2$)$\oplus$($x$-$y$)$^2$$\oplus$($x\cdot y$)). As shown in Table \ref{tab:tab5}, a total of 20 combinations are evaluated on the val set of FIW dataset, where the best result is marked red and the second-best is marked blue. The highest average accuracy is up to $0.716$, when the backbone is ResNet50, loss function is BCE loss, and the operation of feature fusion is ($x^2$-$y^2$)$\oplus$($x$-$y$)$^2$$\oplus$($x\cdot y$).

\subsection{Quantitative Evaluations of Tri-Subject Verification}

\begin{figure}[t]
\setlength{\abovecaptionskip}{0.1cm}
\setlength{\belowcaptionskip}{0.0cm}
\begin{minipage}[b]{1.0\linewidth}
  \centering
  \centerline{\includegraphics[width=8.6cm]{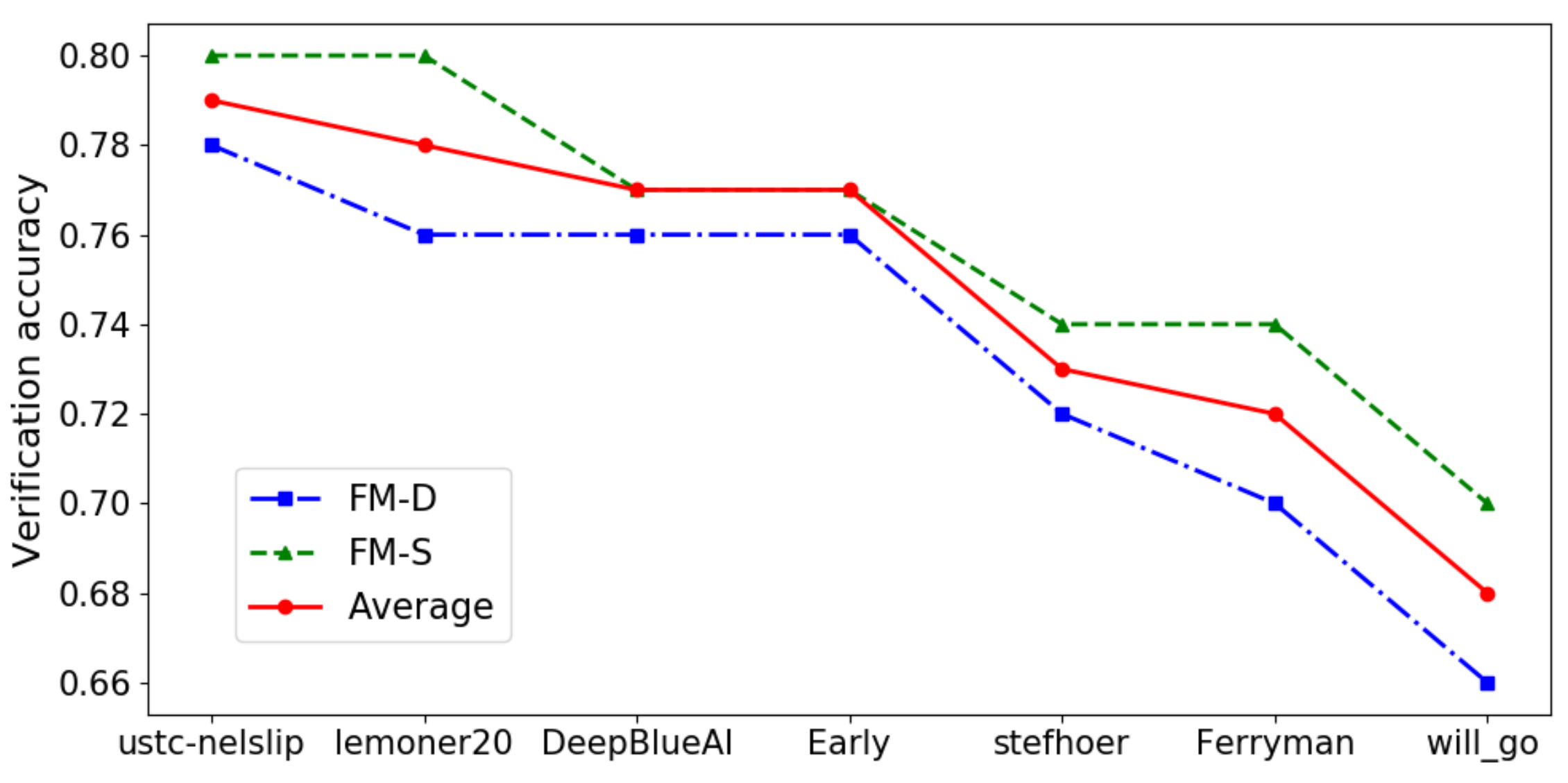}}
\end{minipage}
\caption{Verification accuracy line chart for 2 types of kinship triplets, including top 7 contestants in the leaderboard.}
\label{fig:fig8}
\end{figure}

As illustrated in Table \ref{tab:tab8}, our team (ustc-nelslip) ranks the 1st place in the leaderboard of RFIW2020 Tri-Subject Verification (track II) Challenge, and the average accuracy of our method is up to $0.79$, which is better than that of other contestants. As shown in  Fig. \ref{fig:fig8}, the verification accuracy of FM-S is always higher than that of FM-D in all contestants, and our method is at the highest point in the line chart.

\section{CONCLUSION}

This paper proposes a deep siamese network for bi-subject (1-vs-1) kinship verification and a deep triplet network for tri-subject (2-vs-1) kinship verification. By the operations of feature extraction, feature fusion, and similarity quantification, the deep siamese network is to determine whether two persons belong to the same family. Besides, the deep triplet network is built based on two deep siamese networks, which can quantify the similarity of parent-child kinship, deciding whether a child is related to a pair of parents. In RFIW2020 Challenge, our team (ustc-nelslip) ranked 1st in Tri-Subject Verification track, and 3rd in Kinship Verification track.

\section{ACKNOWLEDGEMENT}

This work was supported by the National Natural Science Foundation of China (NO.U1736123, NO.61572450), USTC Research Funds of the Double First-Class Initiative (YD2350002001).

\end{document}